\documentclass[journal,transmag]{IEEEtran}
\usepackage{graphicx}
\usepackage{graphics}
\usepackage{subfigure}
\usepackage{calligra}
\ifCLASSINFOpdf
\else
\fi
\begin{document}
%
\title{Geometrization of deep networks for the interpretability of deep learning systems}



\author{\IEEEauthorblockN{Xiao Dong, Ling Zhou}
\IEEEauthorblockA{Faculty of Computer Science and Engineering, Southeast University, Nanjing, China}}

%



\IEEEtitleabstractindextext{%
\begin{abstract}
How to understand deep learning systems remains an open problem. In this paper we propose that the answer may lie in the geometrization of deep networks.
Geometrization is a bridge to connect physics, geometry, deep network and quantum computation and this may result in a new scheme to reveal the rule of the physical world.
By comparing the geometry of image matching and deep networks, we show that geometrization of deep networks can be used to understand existing deep learning systems and it may also help to solve the interpretability problem of deep learning systems.
\end{abstract}

}

\maketitle



\IEEEdisplaynontitleabstractindextext

%
\IEEEpeerreviewmaketitle

\section{Motivation}
As a general tool to solve complex problems, the thriving deep learning technology is showing its power in almost all research fields. But we are still lacking a general theoretical framework to to answer the following questions: Why does deep learning work so well? What's the relationship between the structure of a deep network and its functionality? How to design a proper deep network structure for a given task? How can we predict and control the behaviour of a deep network during training? How can our brain construct efficient network structures for different tasks with limited resources?

In this work we propose a general framework to understand deep learning systems, the geometrization of deep networks.

\subsection{Why geometrization}
Our motivation to understand deep learning systems from a geometric perspective falls in three folds.

\textbf{Deep networks are physical}
  The reason that deep learning is so powerful and universally effective in different fields is that deep networks reveal the structures of physical systems. That's to say, deep networks are effective representations of physical systems and their evolutions. Besides the enormous examples of AI based applications on computer vision, natural language processing and robot control, deep networks are also closely related with the fundamental laws of our world, for example the effective representation of many-body quantum systems\cite{Gao2017Efficient}, renormalization group and entanglement renormalization\cite{Evenbly2017Algorithms}\cite{B2013Deep}, tensor networks and AdS/CFT duality\cite{Evenbly2011Tensor}\cite{Swingle2009Entanglement}\cite{Hayden2016Holographic}\cite{Swingle2012Constructing}\cite{Qi2013Exact}\cite{Gan2017Holography}\cite{Dong_deep}. So we believe the effectiveness of deep networks has a fundamental physical origin. That is, the deep network is \emph{at least} a replica of our physical world so that every physical system has a correspondent deep network representation. Deep networks may share the same structure of the physical world and they may obey the same rules. The great success of geometrization of physics inspires our idea that geometrization may also be the ultimate framework to understand deep networks and deep learning systems.


\textbf{Deep networks are computational programmes}
  From the quantum computation point of view, any physical system can be regarded as being generated from an initial simple state by an unitary operation. Similarly the evolution of any physical system is also an unitary operation or equivalently a computation process. As effective descriptions of physical systems and their evolutions, deep networks are essentially computation processes and can be understood as computational programmes to generate and evolve physical systems. Then the geometrization of quantum states and quantum computations\cite{Nielsen_geometry}\cite{Heydari_dynamicdiatance} also leads to the geometrization of deep learning systems. For example, quantum computation complexity has a clear geometric picture and concrete physical meanings as discussed in complexity=action, complexity=volume, Hamiltonian complexity, tensor networks and the emergent spacetime structure from quantum information.

\textbf{Deep networks as optimal control and optimization systems}
  Recently there are emerging efforts to formulate deep learning systems as either optimization or optimal control problems\cite{Tian2018Neural}\cite{Weinan2018A}. It's well-known that these are also closely related with geometry and physics. We will show this point with a concrete example of template image matching, which has a clear geometric picture as an optimization or an optimal control problem.

All the above observations lead to the same conclusion, geometrization scheme may bring us new perspectives to understand deep networks and deep learning systems. What's more, if the geometrization of deep networks can be accomplished, this may also change our ways to understand the physical world, i.e. a physical world built by deep networks.

Now we show how to build the geometrization of deep networks and how this can help us to understand deep networks and deep learning.

\subsection{An abstract description of deep networks}
In order to establish the geometric picture of deep networks, we now give an abstract description of it.

As mentioned above, deep networks are programmes or data processing systems, which can achieve a transformation from the input data space $\mathbf{V}_{in}$ to the output data space $\mathbf{V}_{out}$. Normal deep learning tasks, such as feature extraction and generative models, are all mappings between different data spaces. And usually we prefer one of them to be a vector space so that algebraic operations or classifications can be easily carried out on it. From the general computation point of view, a data processing or a computation system can be abstracted as a mapping $C: V\times G\rightarrow V$, where $V$ is the space of data and $G$ is the space of operations on data. A computation process is given by $C(v_{in},g)=g(v_{in})=v_{out}$, where $v_i,s_o\in V$ are the input and output data and $g\in G$ is an operation or transformation on data. A programme or an algorithm is a realization of $g$, which is usually achieved by a series of simple primitive operations as in both classical and quantum computers. The structure of a deep network is essentially a parametric realization of $g$ and the process of training is to find the proper network parameters that achieve $g$. Here we would like to note that the parameters may also include part of the network structure so that network structure itself may also be learned during training.

The key feature of deep networks, the \emph{deep} structure means that $g$ is realized by a discrete time sequence of transformations $\{\bar{g}_n | n\in[0,N],\bar{g}_0=Id, \bar{g}_N=g\}$, where $Id$ stands for the identity transformation. In this transformation sequence, the n\emph{th} step achieves an operation $\bar{g}_n \circ \bar{g}_{n-1}^{-1}$, which is usually a simple low complexity operation. To make an analytical study of deep learning networks, we introduce a continuous time flow $\{g_t | t\in[0,1],g_0=Id, g_1=g \}$. The validity of this continuous flow fundamentally lies in the continuous evolution of quantum states. This is to say, a discrete time model of quantum information processing system such as the quantum circuit model is essentially only an approximation of a continuous time quantum evolution.  Similarly a discrete time deep network is only an approximation of a continuous flow of transformation.

Obviously the continuous time flow of transformation has a geometric picture. It is a continuous curve in the space of transformation connecting the identity operation $I$ and the target operation $g$. Accordingly for each input data $v_{in}$, there is a curve in the data space given by $\{v_{i,t} | t\in{0,1},v_0=v_{in},v_1=v_{out}\}$. The purpose of deep learning systems is to find an optimal transformation flow to realize the target transformation, where the correspondent collection of the trajectories of all the input data $\{v_{in,t}^k,v_{in}^k \in \mathbf{V}_{in}\}$ should show a good shape, where a good shape means the trajectories should be smooth, stable and well distributed so that the the network has a good genearalization performace.

We now focus on the continuous transformation curve $g_t,t\in[0,1]$. If we regard the space of transformation $G$ as a manifold, then we can build a Riemannian structure on it. We can define the time derivative of $g_t$ as $\dot{g}_t=u_t\circ g_t$ and a right invariant metric on the tangent space $TG$ of G as $<\dot{g}_t,\dot{g}_t>_{TG}=<u_t,u_t>_{Id}$. Then we can calculate the length of the curve $g_t,t\in[0,1],g_0=I,g_1=g$ as $\int_0^1<u_t,u_t>dt$, which is the algorithmic complexity of the realization $g_t,t\in[0,1]$ of $g$.

Now we have a simple geometric picture of deep networks. The structure of a deep network and the metric determines the length of the curve. Network parameters and the metric determines the shape of the curve. The optimal realization of $g$ under a constraint to minimize the length of the curve is the geodesic from $Id$ to $g$.

Of course, keen readers will argue that above geometric picture is too abstract for a quantitative or even a qualitative understanding of deep learning systems. In the remaining part of this paper, we will firstly give a solid example of the geometrization of deep networks by comparing deep networks with the geometry of image matching. Then we scratch a broader picture of the geometrization of deep networks by comparing deep networks with other physical systems including quantum information processing, quantum many-body systems, spacetime structure and general relativity.

\section{Geometry of image registration}

Computational anatomy\cite{Younes2010Shapes} is a research field to study the variability of anatomical shapes, where the comparison between shapes is the key issue. Mathematically a shape can be described by a function on a spatial space $I: R^n\rightarrow R^m$, which we call an image $I$. Here $n=2,3$ stands for a 2D or a 3D image. $m=1$ and $m>1$ mean scalar images and vector/tensor images. For two different shapes represented by correspondent images $I_0,I_1$, the task of image registration is to find a transformation $\varphi$ so that the difference between the target image $I_1$ and the transformed source image $I_0$ is minimized, i.e. $\min_{\varphi}\|I_1-I_0\circ\varphi\|$. The details of the transformation $I_0\circ \varphi$ depends on the type of the image $I_0$\cite{Younes2010Shapes}.

\subsection{Diffeomorphic image registration: optimization vs optimal control}
Diffeomorphic image registration is a framework for shape comparison by modeling transformations between shapes as a smooth invertible function $\varphi:R^n\rightarrow R^n$. For example the space of transformations of volumetric images can be taken as $G=Diff(R^3)$, which is the diffeomorphism group of $R^3$, and $V=I(R^3)$ as the space of volumetric images on $R^3$. Deforming an image $I_0\in V$ by a transformation $\varphi \in G$ is just the change of coordinate as $I_0\circ\varphi$. Following \cite{Bruveris2011The}\cite{Bruveris2013Geometry}, image registration can be abstracted as a map $G\times V\rightarrow V$, where $G$ is the group of diffeomorphic image transformations and $V$ is the vector space of images. Large deformation diffeomorphic metric mapping (LDDMM)\cite{Beg2004Computing} generates a deformation $\varphi$ as a flow $\varphi^u_t$ of a time-dependent vector field $u_t \in T_e(G)=\mathbf{g}$ so that
\begin{equation}\label{eq1}
  \dot{\varphi}^u_t=u_t\circ\varphi^u_{t}, \varphi^u_{0}=Id, \varphi^u_{1}=\varphi
\end{equation}

The diffeomorphic matching of two images $I_0$ and $I_1$ with LDDMM is to find a vector field $u_t, t\in[0,1]$ to minimize the cost function
\begin{equation}\label{eq2}
  E(u_t)=\int_0^1l(u_t)^2dt+\beta|I_1-I_o\circ\varphi^u_{1}|^2, \dot{\varphi}^u_t=u_t\circ\varphi^u_t,\varphi^u_0=Id
\end{equation}
Here the regularity on $u_t$ is a kinetic energy term $l(u_t)=\frac{1}{2}\int_0^1|u_t|^2dt$ with $|u_t|$ a norm on the vector field defined as $|u_t|^2=\langle Lu_t,u_t\rangle_{L^2}$. The operator $L$ is a positive self-adjoint differential operator, for example $Lu_t=u_t-\alpha^2\Delta u_t$. Obviously the norm $|u_t|^2=\langle Lu_t,u_t\rangle_{L^2}$ defines a Riemannian metric on the manifold of the diffeomorphic transformation group $Diff(R^n)$. The second term of $E(u_t)$ computes the difference between the transformed image $I_o\circ\varphi^u_{1}$ and $I_1$.

A necessary condition $DE(u_t)=0$ to minimize the cost function is that the vector field $u_t$ should satisfy the Euler-Poincar\'{e} (E-P) equation
\begin{equation}\label{eq3}
  Lu_t=-\varphi^u_{0,t}I_0\diamond \varphi^u_{0,t}\varphi^u_{1,0}\pi
\end{equation}
where $\varphi^u_{s,t}=\varphi^u_t\circ\varphi^u_{s^{-1}}$, $\pi:=\beta(\varphi^u_{0,t}I_0-I1)^\flat \in V^*$. The $\flat$ operator is defined as $\flat:V\rightarrow V^*,\langle u^{\flat},v\rangle_{V^*\times V}=\langle u,v\rangle$ and $\diamond:TV^*\rightarrow \mathbf{g}^*,\langle I\diamond \pi,u\rangle_{\mathbf{g}^*\times \mathbf{g}}=\langle\pi,\zeta_u(I)\rangle_{V^*\times V}$ is the momentum map.

The E-P equation can also be given as
\begin{equation}\label{eq4}
  \frac{d}{dt}\frac{\partial l(u_t)}{\partial u_t}=-ad^*_{u_t}\frac{\partial l(u_t)}{\partial u_t}
\end{equation}
where $\frac{\partial l(u_t)}{\partial u_t}$ is the momentum and $ad^*:\mathbf{g}\rightarrow gl(\mathbf{g})$ is the coadjoint representation of the Lie algebra $\mathbf{g}$ of the Lie group $G$. For more details please refer to \cite{Bruveris2011The}\cite{Bruveris2013Geometry}.

In LDDMM framework, the curve satisfying the E-P equation is found by a gradient descent algorithm, while the gradient is given by $u_t+K\varphi^u_{0,t}I_0\diamond \varphi^u_{0,t}\varphi^u_{1,0}\pi$ with $K=L^{-1}$.

The geometric picture of LDDMM is quite simple: LDDMM finds a minimal length curve, i.e. a geodesic given by the E-P equation, in $Diff(R^n)$ connecting $Id$ and $\psi$, which can transform the source $I_0$ to a near neighbour of the target image $I_1$. Equivalently we can also induce a Riemannian structure on the image space $V$ by the map $G\times V\rightarrow V$ so that the geodesic on G leads to a geodesic on V\cite{Bruveris2013Geometry}.

Here we point out that this is exactly the same as in the geometry of quantum computation\cite{Nielsen_geometry}\cite{Nielsen_geometry2} that a Riemannian metric on the quantum operation group induces a Riemannian metric on the Hilbert space of quantum states. Another interesting observation is that the map $G\times V\rightarrow V$ can also be understood as a typical computation system, where $V$ is the data representation space and $G$ is the data operation space. So in fact image registration and quantum computation essentially have the same abstract descriptions and geometric pictures\cite{Dong_deep}.

LDDMM based image registration is formulated as an optimization problem and solved by a gradient descent based optimization. The optimal solution $\varphi_t$ is parameterized by the time-dependent vector field $u_t$ and the optimization procedure is a parameter estimation of $u_t$. We can easily see this is very similar with the abstract model of deep networks we introduced above.

An alternative framework of LDDMM is to formulate it as an optimal control problem\cite{Hart2013An}, where the image registration procedure is regarded as a dynamical process. The state of the dynamical system is the transformed source image $I_0\circ\varphi_t$ and the vector field $u_t$ is taken as the control signal to adjust the transformation $\varphi_t$. The problem is then to minimize the energy function

\begin{eqnarray}\label{eq5}
  E(u_t,J_t^0,\lambda_t,\gamma) &=&\int_0^1l(u_t)+\langle\lambda_t,\dot{J}_t^0+\nabla I_t\cdot u_t\rangle dt \\
   &+& <\gamma,J_0^0-I_0>+\beta|J_1^0-I_1|
\end{eqnarray}
where $J_t^0=I_0\circ\varphi^u_{0,t}$, $J_t^1=I_1\circ\varphi^u_{1,t}$ and $\lambda_t,\gamma$ are the Lagrangian multipliers.

This leads to the optimality conditions as follows
\begin{eqnarray}\label{eq6}
  \dot{J}_t^0+\nabla J_t^0\cdot u_t &=& 0 \\
  \dot{\lambda}_t+\nabla\cdot(\lambda_t\cdot u_t) &=& 0 \\
  u_t+K\star\nabla J_t^0\lambda_t &=& 0 \\
  J_0^0 &=& I_0 \\
  \lambda_1 &=& \beta(I1-J_1^0)
\end{eqnarray}

The optimization procedure is a bi-directional information flow. Given the current control signal $u_t$ and initial values of $J_0^0=I_0$, the forward information flow compute $J_t^0$ for $t\in[0,1]$. In the backward adjoint flow, we update $\lambda_t$ starting from $\lambda_1=\beta(I1-J_1^0)$ and then $u_t$ can be updated by a gradient descent using both $J_t^0$ and the adjoint variable $\lambda_t$.

We note that the gradient based update of $u_t$ here is in fact the same as the updating of $u_t$  in the optimization formulation to fulfill the E-P equation. But the idea of bi-directional adjoint computation is a new characteristic. This is different from the direct computation of gradient in the optimization formulation. The Lagrangian multiplier based formulation can lead to more general strategies for parameter optimization as will be shown later.

\subsection{Geodesic shooting}
In LDDMM, both the optimization and optimal control formulations aim to find a geodesic by finding a vector field $u_t$ satisfying the E-P equation. It's well known that for a given Riemannian manifold, a geodesic is completely determined by the starting point and the initial velocity of the geodesic. So if our goal is to find a geodesic, then the vector field $u_t$ as a control signal is highly redundant since it can be completely determined by $u_0$ and the E-P equation.

Geodesic shooting\cite{Vialard2012Diffeomorphic} can find the initial vector field $u_0$ or equivalently the initial momentum $Lu_0$ with the E-P equation as an explicit constraint. Obviously here the geodesic shooting is also formulated as an optimal control problem. The correspondent optimization procedure is also a bi-directional information flow. Starting from the initial momentum and the E-P equation, the forward flow updates the vector field $u_t$, the transformation $\varphi^u_t$ and the transformed source image $J_t^0=I_0\circ\varphi^u_t$. The backward adjoint flow updates the adjoint variables, the Lagrangian multipliers of the constraints, and finally the initial vector field $u_0$ can be updated by a gradient descent. For more details of geodesic shooting and the related adjoint calculation, please refer to \cite{Vialard2012Diffeomorphic}\cite{Younes2010Shapes}.

The lesson we can draw from geodesic shooting is that, when the optimal configuration of a subset of the parameters can be determined as a function of all the other parameters, this function can be regarded as a constraint and the optimization can be simplified as an optimal problem. Or in another word, when there exists explicit constraints among parameters, the optimization can be simplified. This may help to design the network structure in deep learning systems.

This idea can be generalized to the case of a general optimal control, where explicit constraints among parameters should be respected. Then the Lagrangian multiplier based variational method will lead to a similar bi-directional information passing algorithm which can be shown later to be closely related with deep learning systems.

\subsection{Semiproduct group and metamorphosis}
Metamorphosis is used to modify the original LDDMM or the geodesic shooting to support a second transformation flow $\eta_t, v_t=\varphi^u_t\dot{\eta}_t$, which is used to change the image appearance of the template image $I_0$ so that the image transformation is a composition of both the coordinate transformation and the image appearance transformation. Essentially this is to replace the Lie group $G$ with a semiproduct group\cite{Bruveris2013Geometry}\cite{Holm2008The}. That's to say, we are now working with a composite operation of multiple operations.

Under the composite transformation, the image is transformed as $\dot{J}_t^0=v_t+u_t\circ \varphi^u_t$. This is a new constraint involving both the coordinate and image appearance transformations. Accordingly the energy function to be minimized includes both the kinetic energies of $u_t$ and $v_t$. In another word, the Riemannian manifold of transformations is extended and a new composite metric is defined. But basically the geometric picture is similar with the original LDDMM or the geodesic shooting framework. An alternative perspective of the metamorphosis is that the transformation on the image appearance $\eta_t$ can be regarded as introducing noise on the image. The constraint on the kinetic energy of $v_t$ can be understood as to constrain the power of the noise. For more details of the idea of metamorphesis, please refer to \cite{Younes2010Shapes}\cite{Holm2008The}\cite{Holm2009Geometric}.

\subsection{Summary of diffeomorphic image registration}
Image registration can be formalized by either energy based optimization or an optimal control problem. It has a clear geometric picture, where the optimal solution is a geodesic on the Riemannian manifold on the transformation space. The geodesic is represented by a parameterized model and is obtained by a parameter estimation procedure. What's more, the image registration problem is closely related with the geometric mechanics in that they share lots of geometric structures.

A complete description of the geometric structure of image registration is given in \cite{Bruveris2011The}\cite{Bruveris2013Geometry}\cite{Vialard2012Diffeomorphic}\cite{Holm2008The}\cite{Holm2009Geometric}\cite{Younes2010Shapes} and references therein.


\section{Geometric picture of deep learning systems}
To show the validness of the geometrization of deep networks, we now compare the diffeomorphic image registration and deep learning systems to build a dictionary between correspondent concepts in both fields. Since image registration has both a clear geometric and a physical (geometric mechanical) picture, we hope the dictionary will give us a new understanding of deep networks from both the geometric and physical points of view. Here we directly give a list of the content of the dictionary with a brief explaination. Interested readers can check the details by themselves.

\subsection{A dictionary between image registration and deep networks}
(1)\textbf{Network structure and $G$}: Geometrically the network structure defines the space of possible solutions, which is a set of curves in the transformation space. Also the network structure defines in which way this space is explored as explained in the relational inductive bias\cite{DBLP:journals/corr/abs-1806-01261}. Due to the limited complexity of the network and limited allowed operations, the network structure only represents a subset of all possible curves that can reach the target transformation $\varphi$ in image registration or $g$ in deep learning from the identical transformation $Id$. In fact the deep network structure defines the operation group $G$ and the network parameters $\theta$ falls in its Lie algebra $\mathbf{g}$ of $G$. Normal CNNs are just discretisized transformation curves and the norms of network parameters $\theta$ along the curve can be roughly regarded as the non-uniform discretization step sizes.

(2)\textbf{Constraints and Riemannian metric}: The network structure only defines the space of possible solutions. To find the optimal solution by solving an optimization problem, we need to introduce constraints on the parameters of the network. Geometrically constraints can be regarded as Riemannian metrics on $G$ defined on the manifold of possible solutions encoded in the network structure and network parameters. Carefully adjusting constraints can change the curvature distribution of the solution manifold and generally we prefer to work on a flat manifold so that the optimal solution can be easily found.

(3)\textbf{Supervised training and landmark registration}: Given the parametric description of the manifold of possible solutions and the Riemannian metric defined by constraint, supervised training on a set of $N$ labeled training data estimates the parameters $u_t$($\theta$) to find the optimal transformation curve $g_t$ to reach the desired target transformation. This can be understood to achieve a diffeomorphic image registration based on $N$ pairs of landmarks on the image or to simultaneously match $N$ images using the same diffeomorphic transformation.

(4)\textbf{Optimal deep networks and geodesics}: In LDDMM, the optimal transformation is achieved by a geodesic on $G$ determined by the E-P equation. In deep networks, an optimal network should also exist as a geodesic on the Riemannian manifold defined by the network structure and constraints, which is the deep network with a minimal complexity. Accordingly, if the E-P equation of deep networks can be explicitly described as in LDDMM, then geodesic shooting can also be implemented on the optimization of deep networks. Since geodesic shooting only optimize the initial momentum of the geodesic, the training of an optimal deep network has a much smaller degree of freedom than a normal non-optimal deep network. Network distillation and network pruning are essentially both efforts to find the optimal deep networks.

(5)\textbf{Back propagation and LDDMM}: The back propagation based optimization of deep networks are essentially the same as the gradient descent based LDDMM optimization\cite{Beg2004Computing}.

(6)\textbf{Neural ODE and optimal control framework}: The optimal control based optimization procedure of LDDMM is essentially the same as the optimization used in neural ODE and other related works\cite{Tian2018Neural}\cite{Weinan2018A}.

(7)\textbf{Equilibrium propagation and geodesic shooting}: Bengio's equilibrium propagation\cite{Scellier2017Equilibrium} share the same structure as geodesic shooting used in LDDMM framework\cite{Vialard2012Diffeomorphic}.

(8)\textbf{Attention mechanism and semiproduct group}: Essentially attention mechanism is a composition of multiple deep networks. It shares lots of similarity with the semiproduct group and metamorphoses in LDDMM framework since semiproduct group also plays with composite operations. In the semiproduct group case of LDDMM, the multiple operations are coupled and a generalized E-P equation can obtained to represent the optimal flow. This indicates that theoretically we also have an optimal attention mechanism and the geodesic shooting scheme can be applied.

(9)\textbf{Generalization and Riemannian curve length}: The generalization capability of deep networks is a key issue of the performance of deep networks. Usually generalization is described by a norm based factor, which can be understood as the complexity of the network\cite{Liang2017Fisher}. It has been found that deep networks have the tendency to reduce complexity during training and a lower complexity means a better generalization capability. In LDDMM, the complexity of the registration transformation is the length of the transformation curve evaluated using the Riemannian metric on $G$. In deep networks, we also have a correspondent network complexity using a special Riemannian metric, the Fisher-Rao metric\cite{Liang2017Fisher}. In fact this metric is closely related with general relativity\cite{Matsueda2013Emergent}\cite{Matsueda2014Derivation}\cite{Matsueda2014Geodesic}, which is another evidence that deep networks have a deep physical origin. We will give more details on this point in the discussion section.

(10) \textbf{Batch normalization and geodesics}: It's well known that batch normalization can help the convergence of deep networks. From the geometrical point of view, since CNNs are just discretisized transformaition curves and the norms of $\theta$ are the discretization step sizes, BN can be regarded as an operation to adaptively adjust the ratio of thrown-away information (energy) along the curve. This is because CNNs are the same as the entanglement renormalization algorithm, which extracts global information by iteratively throwing away local information\cite{Evenbly2017Algorithms}. By normalizing the data, BN aims to keep a constant speed of throwing away information along the network. In entanglement renormalization algorithms, this is to throw away a fixed percentage of low amplitude states and keep only those high amplitude states so that the strong global information patterns are kept. This will result in a transformation curve with an isometric-like property, which coincides the property of geodesics. So geometrically BN can be understood as a constraint to force the network to be a geodesic-alike curve. This geometric picture is the same as the conclusion of \cite{NIPS2018_7515}. In \cite{NIPS2018_7515} the Hessian matrix was introduced, which is in fact the Fisher-Rao metric to evaluate the complexity of the deep network or the length of the transformation curve. If the network can be constrained by BN to form a geodesic-like curve, then it has the minimal curve length or equivalently the minimal deformation energy as in the geometry of image registration problem. Or BN forces the Fisher-Rao metric to vary smoothly along the network. Obviously a transformation with a minimal deformation energy or a smooth curve will have a smooth loss landscape, which coincides with a key conclusion of \cite{NIPS2018_7515}.

(11)\textbf{Training convergence and curvature}: The convergence of deep networks highly depends on the back propagation of gradients along deep networks. From the geometric picture of LDDMM, we know that this is related with the curvature of the manifold since the curvature determines the stability of geodesics. In deep learning fields, random matrix based analysis\cite{Xiao2018Dynamical} shows that when the network has a dynamic isometric property, the forward and backward information can flow freely along the network so that a better convergence can be achieved. In fact, isometry is exactly the property of a geodesic. So the dynamic isometric property of a deep network is essentially to say, the network is a geodesic on Euclidean manifold, i.e. a straight line. Similarly, batch normalization is essentially to adjust the curvature of the manifold by adjusting the Fisher-Rao metric along the network.

(12)\textbf{GAN and current based shape matching}: The key goal of GAN is to approximate a distribution density. The main challenge of GAN is to find a proper metric to measure the difference of distributions. This is why WGAN emerges as a break-through since it provides an efficient metric for distributions without one-to-one correspondence between samples of distributions. In LDDMM, there is also a way to compare two shapes without position correspondence using current based shape representation\cite{Durrleman2009Statistical}. It will be interesting to find if there exists a correspondence between WGAN and currents.

(13)\textbf{Dropout and stachastic shape evolutions}: As a solution to enhance the robustness of deep networks, dropout achieves its goal by adding perturbations on the network, either on the operation group $G$ (dropout of neurons) or on the data space $V$ by adding perturbation layers\cite{You2018Adversarial}. In LDDMM framework, there are also similar shape registration methods by adding perturbations on either the momentum or the positions of landmarks on shapes\cite{Trouv2012Shape}. Obviously dropout aims to find a curve that is robust against perturbations on either \textbf{G} or \textbf{V}. But how about a perturbation on the Lie algebra $\mathbf{g}$? We will also address this issue in the following section.

(14)\textbf{ResNet, Lie algebra,curvature and reparameterization of curves}: ResNet as the most successful deep network structure shows a superior performance than normal CNNs. From a geometric point of view, the success of ResNet falls in that ResNet is a network running on the Lie algebra $\mathbf{g}$, while normal CNNs run on $G$. This can be understood by taking the quantum computation as an analogue of CNNs\cite{Dong_deep}, where normal CNNs try to construct an quantum algorithm by composing elementary unitary gates and ResNets achieve the same algorithm by finding the proper Hamiltonian. It's well known that ResNet is essentially a differential equation, which perfectly matches the structure of LDDMM. For ResNets, the curvature along the network is much smoother than normal CNNs since the network parameters of ResNets are only weak perturbations and therefore can not lead to rough curvature change along the network. ResNets can also easily achieve reparameterization of the curve $g_t$ by just adjusting the amplitudes of the weights. In another word, compared with normal CNNs, ResNets run on a much smoother manifold and can approach a smooth geodesic much easier than normal CNNs.

(15) \textbf{Geometric structure of deep networks and Riemannian structure on $V$}: In deep learning fields, there are also works to explore the geometric structure of deep networks\cite{NIPS2016_6322}\cite{NIPS2017_6873}. These works are closely related with the geometrization of deep networks. But both of them are working on the Riemannian geometry on $V$ as described in \cite{Bruveris2013Geometry} instead of on the Riemannian geometry on $G$. Since the geometry on $V$ is induced by a projection from the geometry of $G$, a complete geometrization of deep networks should be accomplished on $G$ and only the geometry of $G$ can fully explore the dynamics of deep networks.

This is only a partial list of the correspondence between the geometry of image registration and deep learning systems. We hope we have convinced readers to believe the geometrization of deep networks is a promising candidate for the interpretation of deep learning systems.

\subsection{A concrete example}
Now we will give a concrete sample on how we can understand deep learning using the geometrization framework. In \cite{pmlr-v70-koh17a} the problem of how the training data will influence the prediction of a deep network was addressed. They considered a supervised training with $n$ data points $z_i={x_i,y_i}, i=1....n$ and the cost function is $L(z,\theta)\frac{1}{n}\Sigma_{i=1}^nL(z_i,\theta)$ with $\theta$ as the network parameters. The optimal network configuration is given by $\hat{theta}=\arg\min\min_{theta}L(z,\theta)$.

The key results of \cite{pmlr-v70-koh17a} are two items to evaluate how the perturbation on the training data will influence the parameter $\hat{theta}$ and the loss at a test point $z_{test}$ given by
\begin{equation}\label{eq7}
  I_{up,params}(z)=-H_{\hat{\theta}}^{-1}\nabla_\theta L(z,\hat{\theta})
\end{equation}
\begin{equation}\label{eq8}
  I_{up,loss}(z,z_{test})=-\nabla_\theta L(z_{test},\hat{\theta})H_{\hat{\theta}}^{-1}\nabla_\theta L(z,\hat{\theta})
\end{equation}
where $H_{\hat{\theta}}=\frac{1}{n}\sum_{i=1}^n\nabla_{\hat{\theta}}^2L(z_i,\hat{\theta})$ is the Hessian.

To interpret the results using our geometrization framework, we can regard the supervised network training as an optimization problem following the formulation of image registration with a cost function
\begin{equation}\label{eq9}
  E(u_t)=\int_0^1l(\theta)dt+L(z,\theta)
\end{equation}
where $l(\theta)=\langle\theta\mathbf{I}(\theta)\theta\rangle$, $\mathbf{I}(\theta)=\sum_{i=1}^n[\nabla_{\theta}L(z_i,\theta)\otimes\nabla_{\theta}L(z_i,\theta)]$ is the Fisher-Rao metric used in \cite{Liang2017Fisher} to describe complexity of deep networks. Obviously the Fisher-Rao metric is essentially the same as the Hessian $H_{\hat{\theta}}$ since $\mathbf{I}(\theta)=-H_{\hat{\theta}}$ from information geometry.

This can be understood as either to match $n$ pairs of images simultaneously using diffeomorphic transformations on a higher dimensional space (due to the overparameterization of deep networks) or a landmark based image registration taking all the training data as paired landmarks on an image. The goal of the optimization is to find a proper transformation that can match the training data and also show good generalization performance. The Riemannian metric on the Riemannian manifold to measure the deformation energy of the transformation or the curve length or equivalently the complexity of the deep network is the Fisher-Rao metric of the deep network. From a physical point of view, it's obvious to see that why the Fisher-Rao norm is used in \cite{Liang2017Fisher} to represent the generalization capability. This is because a lower complexity network means a lower deformation energy and therefore a smoother image deformation field. Of course for a landmark based image registration, a smooth deformation will have a better generalization performance.

Comparing (\ref{eq9}) with (\ref{eq2})(\ref{eq3})(\ref{eq4}), we can observe that $\nabla_\theta L(z,\hat{\theta})$ in (\ref{eq7}) is exactly the momentum in $\mathbf{g}^*$ of E-P equation and (\ref{eq7}) is the correspondent vector in $\mathbf{\mathbf{g}}$. \ref{eq8} is related with the angle between two vectors in $\mathbf{g}$ using the Fisher-Rao metric. We can easily draw a clear physical or a geometric picture of (\ref{eq7})(\ref{eq8}). If the deep network is a geodesic under the Fisher-Rao metric, then roughly (\ref{eq7}) indicates how the direction of the geodesic will be shifted with a perturbation of a landmark. (\ref{eq8}) shows under a perturbation of a training landmark, how the direction of the trajectory of a test data $z_{test}$ transformed by the perturbed geodesic will be shifted. Of course generally deep networks are not geodesics. Then the networks in \cite{pmlr-v70-koh17a} are just normal non-geodesic curves. But still the above geometric picture holds approximately.

The drawback of \cite{pmlr-v70-koh17a} is that it only consider a perturbation around the current configuration $\hat{\hat{\theta}}$ so that the Fisher-Rao metric is fixed by the network configuration. Another more interesting work \cite{pmlr-v80-ren18a} tried to explore the complete dynamics of the Fisher-Rao metric by iteratively updating the weighting of training data and the network configuration $\hat{\theta}$. Similar to \cite{pmlr-v70-koh17a} this can be formulated as an image registration problem give by

\begin{equation}\label{eq10}
  E(u_t)=\int_0^1l(\hat{\theta(\epsilon)})dt+L(z,\theta,)
\end{equation}
with $\epsilon_i,i=1,...,n$ are the weights of the training data and $l(\hat{\theta(\epsilon)})$ are defined on the Fisher-Rao metric determined by the optimal network parameters $\hat{\theta(\epsilon)}$ which are $\epsilon$ dependent.

What's new here? The main difference with the image registration problem is that the Fisher-Rao metric on $G$ is now data dependent! In another word, the Riemannian metric is not a fixed background metric as in the image registration problem, instead the metric is emergent from the deep network itself. Readers with a physics background can immediately see that we have an analogue of this in physics. The data independent image registration is the Neutonian mechanics with a fixed spacetime background and the data dependent deep network systems correspond to general relativity with a dynamic spacetime. What's more, the Fisher-Rao metric used here is in fact closely related with general relativity since gravitation equation can be derived from it\cite{Matsueda2013Emergent}\cite{Matsueda2014Derivation}\cite{Matsueda2014Geodesic}. So in (\ref{eq10}) the network structure and data (information) are coupled just as spacetime and matter are coupled in general relativity. \emph{Following John Wheeler, in our physical world, spacetime tells matter how to move, matter tells spacetime how to curve. In deep networks, network tells data (information) how to move, data (information) tells network how to curve.} We believe this is not just an analogue between the physical world and deep networks, this should be regarded as a general principle to design and understand deep networks. The key component of interpret deep networks is to understand how the network structure and data information interact. That's to say to find the gravitation equation for deep networks. Here we point out, since the Fisher-Rao metric is data dependent, the optimal solution can not be written as the E-P equation with a fixed Riemannian metric any more since the metric is also dynamic.  In \cite{pmlr-v80-ren18a}, the solution is approximated by a two-level gradient descent algorithm which updates the network parameter $\hat{\theta}$ and sample weights $\epsilon$ iteratively. The final solution is a critical point that $\hat{\theta}$ is stable with respect to the perturbation of $\epsilon$. At the critical point, the solution still satisfies the E-P equation with a Fisher-Rao metric determined by the network parameter $\hat{\theta}$.

As a conclusion of this section, the geometrization framework tells us that (1)The Fisher-Rao based network complexity measure is an effective signature for the generalization property for deep networks since it's a measure for the network complexity; (2)The network structure and data information are coupled just as matter and spacetime are coupled in general relativity and the ultimate law of deep networks is a gravitational equation of deep networks; (3)The optimal solution is a result of the competition between the two terms, the network complexity and training error, in (\ref{eq10}).

\section{Discussion}
Till now we have seen the validness of the geometrization framework on the interpretability of deep learning systems by showing that deep networks can correspond to geometrical mechanics and general relativity. The basic idea of geometrization is that deep networks have correspondence in the physical world. Therefore we can regard deep networks as physical systems and ask the following questions:

(1) Is there a GUT (grand unified theory) of deep networks?
If there is a correspondence between deep networks and our physical world, then the ultimate interpretability of deep networks lies in finding the GUT of deep networks just as the interpretability of the physical world lies in the GUT of the physical world. It's a common sense that the physical GUT is definitely a geometrical theory. So we believe geometrization should be the right roadmap for the interpretability problem of deep learning systems. Also the GUT of deep networks should have the same structure as the GUT of our physical world. Therefore even the GUT of our physical world is not available yet, exploring the similarity between physical systems and deep networks can provide guidelines for us to better understand deep networks. 

(2) Real physical systems obey a least action principle, is this also true for deep networks?
We have seen that the geometry of image registration results in an optimal solution given by the E-P equation. But for deep networks, generally we are working with systems far from optimal. But still usually these non-optimal systems work well in practice. There are also works taking deep networks as general dynamic systems such as in neural ODE\cite{Tian2018Neural}. Shall we investigate non-optimal deep networks as general information processing systems or should we stick to optimal deep networks since they are more physical? Our geometrization framework will definitely work better on the optimal deep networks. But non-optimal systems might not be properly geometrized. So maybe we should first focus on understanding the optimal systems with clear geometric pictures. 

(3) What can we learn from the geometrization of physics?
Till now we are only working with Riemannian structures as in the geometry of image registration. In fact the geometrization of physics is far beyond Riemannian structures. A natural extension is the fibre bundle structure which plays a key role in gauge theory. Can we describe deep networks using fibre bundles? Good candidates in deep networks that may be described by fibre bundles are transfer learning, meta learning, neural Turing machines (NTM) and differentiable neural computers(DNC). They all aim to find some kind of \emph{reconfigurable} systems. In the language of Riemannian geometry, this usually means to reconfigure the metric of the system so that the optimal geodesic curves can be reconfigurable. A natural way to achieve this is to reconfigure the connection form on fibre bundles. Roughly transfer learning can be understood as to transfer (part of) a geodesic to another task. Meta learning aims to find some universal descriptions of different but similar geodesics. NTM and DNC mainly achieve the refiguration of systems by carefully changing their memories. We can see the memory can be understood as the fibre bundle above the base space of the LSTM states. It's interesting to check if NTM and DNC can be written as defining a connection on their fibre bundles.

Another possibility is that fibre bundles can be used to describe the coupling of multiple deep networks. It's getting more and more obvious that complex tasks can only be achieved by coupling multiple deep networks just as in our human brains. In AI systems, typical coupled composite systems are GANs and attention. The coupling of systems leads to interactions between subsystems, just as interactions (forces) between physical systems. In physics, interactions are described by fibre bundles. Accordingly interactions between coupled deep networks should also be described by fibre bundles.

The last but not the least, the coupling of multiple deep networks might be related with the existence of consciousness. We hypothesize that when multiple neural networks in our brains are coupled, the coupling may be achieved by an independent coupling system, which not only couples the multiple neural subsystems but also has its own latent state space and a stable dynamics. This independent coupling system may be the origin of our consciousness. If this is the case, then can our consciousness also be geometrized?

(4) How to geometrize reinforcement (RF) learning systems? Geometrically RF is essentially to learn the metric of $G$ from the interaction with the system and then find geodesics using the learned metric. Imitation learning can be understood to design a metric so that the expert's action becomes a geodesic. Can we formulate a geometrization of these procedures?

(5) What's the curvature of the emergent Fisher-Rao metric in deep networks?
If deep networks can be formulated as a dynamic system using emergent Fisher-Rao metric, we need to check what's the curvature of this metric. Because the curvature will determine the stability of the geodesic. And the metric is dependent on both the structure and the parameters of the network. Just as in general relativity, the solution spacetime can have either positive or negative curvature, deep networks may have the same problem. Taking CNN as an example, in quantum information field the correspondent system is MERA or the entanglement renormalization algorithm which show a similar structure as CNN. We know that MERA builds a negative curvature geometry and is related with the famous AdS/CFT duality. Similarly the geometry of quantum computation, which is another analogues of image registration and CNNs also has an almost negative curvature\cite{Nielsen_geometry2}, where the Riemannian metric used here is static just as in image registration. It's reasonable to guess that CNN may also have a similar negative curvature. This might be an explaination of the existence of adversarial examples in CNN based classification networks.

(6) How to understand the overparameterization of deep networks?
Overparameterization plays a key role in nowadays deep networks. It's closely related with the training convergence, generalization and adversarial attacks. Geometrically this means to choose a higher dimensional group $G$ to accomplish the transformation. In physics we also meet overparameterization problems. For example, in quantum computation overparameterization means to achieve a quantum algorithm using auxiliary qubits\cite{Nielsen_geometry2}. In tensor network representation of quantum states, overparameterization is closely related with the concepts of parent and uncle Hamiltonians\cite{Fern2015Frustration}. Overparameterization not only brings a higher dimensional $G$ but also a potentially more flexible network structure. As we see above, the structure of deep networks will influence the curvature of the geometry built by the Fisher-Rao metric of networks. A complete understanding of the consequence of overparameterization is still needed.

\section{Conclusions}
In this work, inspired by the geometrization of physics, we proposed a geometrization framework for the interpretability of deep learning systems. By comparing the geometry of image registration with deep networks, we showed that geometrization does bring us new pictures of deep networks. Under this framework, we also discussed some key problems for the understanding of deep learning systems. Our future work will be then to answer these questions.

As a final remark, besides the geometrization of physics to connect physics and geometry, currently there is a trend to understand physical laws from the computation point of view so that computational complexity starts to play a key role in physics. If we further bring deep networks into this game, we hope the interactions among physics, geometry, computation and deep networks may completely change our understanding of the world. A possible picture of our world may be: \textbf{\emph{The world is an information processing (computation) system that generates our universe by a deep network of basic computational operators. The structure of the deep network is determined by the information structure of our universe. That's to say the deep network is the optimal network to generate the information pattern of our universe, i.e. a geodesic according to a certain Riemannian metric to measure the computational complexity. Physical laws are encoded in the correspondence between the geometric structure of the network and the information pattern of our universe. So still our world obeys a least action principle with the action is given by the computational complexity of the physical world.}} 

\bibliographystyle{unsrt}

\bibliography{DLgeometrization}





\end{document}